%% file: output.tex
\documentclass[letterpaper, 10 pt, conference]{ieeeconf}  

\IEEEoverridecommandlockouts                              

\makeatletter
\let\NAT@parse\undefined
\makeatother
\input{macro}

\input{00_title_authors}

\begin{document}

\maketitle
\thispagestyle{empty}
\pagestyle{empty}
\renewcommand{\baselinestretch}{0.979} 

\input{01_abstract}

\renewcommand{\baselinestretch}{0.979}

\input{1_intro}

\renewcommand{\baselinestretch}{0.979} 
\input{2_prior}

\renewcommand{\baselinestretch}{0.979} 
\input{3_method}
\renewcommand{\baselinestretch}{0.979} 
\input{4_exper}

\renewcommand{\baselinestretch}{0.979} 
\input{5_disc}
\renewcommand{\baselinestretch}{0.979} 
\input{6_ackn}


\addtolength{\textheight}{-0.2cm}   


\bibliographystyle{ieeetr}

\bibliography{output.bbl}

\newpage

\input{7_appdix}
\end{document}

%% file: macro.tex
\usepackage{graphics} 
\usepackage{graphicx}
\usepackage{grffile} 
\usepackage{amsmath} 
\usepackage{amssymb}  
\usepackage{color}
\usepackage{mathtools} 
\usepackage{booktabs,ragged2e}
\usepackage{breqn}
\usepackage{textgreek}
\usepackage[flushleft]{threeparttable}
\usepackage{tabularx,booktabs}
\usepackage{graphics}
\usepackage{amsmath}
\usepackage{gensymb}
\usepackage[normalem]{ulem} 
\usepackage{pifont}
\usepackage{booktabs}
\usepackage{multirow}

\usepackage{lineno}
\usepackage{cite}
\usepackage{bm}
\usepackage{url}

\usepackage[noend]{algpseudocode}
\usepackage{algorithm}
\usepackage{makecell}
\usepackage[square, comma, numbers,sort&compress]{natbib}
\usepackage{multirow}
\let\labelindent\relax
\usepackage{enumitem}
\usepackage[table,dvipsnames]{xcolor}
\usepackage[pagebackref=true,breaklinks=true,colorlinks,bookmarks=false]{hyperref}
\usepackage{comment}

\hypersetup{
linkcolor=BrickRed
,citecolor=Green
,filecolor=Mulberry
,urlcolor=NavyBlue
,menucolor=BrickRed
,runcolor=Mulberry
,linkbordercolor=BrickRed
,citebordercolor=Green
,filebordercolor=Mulberry
,urlbordercolor=NavyBlue
,menubordercolor=BrickRed
,runbordercolor=Mulberry
}

\usepackage[disable]{todonotes}
\usepackage[normalem]{ulem} 

\newcommand{\wenzhen}[1]{\todo[inline,color=red!40]{Wenzhen: #1}}

%% file: 00_title_authors.tex
\title{\LARGE \bf
Toward Zero-Shot Sim-to-Real Transfer Learning \\for Pneumatic Soft Robot 3D Proprioceptive Sensing
}

\author{\small Uksang Yoo$^{1*}$, Hanwen Zhao$^{2*}$, Alvaro Altamirano$^{2}$, Wenzhen Yuan\textsuperscript{1,\ding{41}}, Chen Feng\textsuperscript{2,\ding{41}}
\thanks{$^{*}$ indicates equal contributions.}%
\thanks{\ding{41} Corresponding authors. The work is supported by NSF grant 2024882 and NSF Graduate Research Fellowship under Grant No. DGE2140739.}%
\thanks{$^{1}$Robotics Institute, Carnegie Mellon University, Pittsburgh, PA 15213, USA
	{\tt\small \{uyoo, wenzheny\}@andrew.cmu.edu}}%
\thanks{$^{2}$New York University, Brooklyn, NY 11201, USA
	{\tt\small \{hzhao, aa9420, cfeng\}@nyu.edu}}%
}

%% file: 01_abstract.tex
\begin{abstract}
Pneumatic soft robots present many advantages in manipulation tasks. Notably, their inherent compliance makes them safe and reliable in unstructured and fragile environments. However, full-body shape sensing for pneumatic soft robots is challenging because of their high degrees of freedom and complex deformation behaviors. Vision-based proprioception sensing methods relying on embedded cameras and deep learning provide a good solution to proprioception sensing by extracting the full-body shape information from the high-dimensional sensing data. But the current training data collection process makes it difficult for many applications. To address this challenge, we propose and demonstrate a robust sim-to-real pipeline that allows the collection of the soft robot's shape information in high-fidelity point cloud representation. The model trained on simulated data was evaluated with real internal camera images. The results show that the model performed with averaged Chamfer distance of $8.85$ mm and tip position error of $10.12$ mm even with external perturbation for a pneumatic soft robot with a length of $100.0$ mm. We also demonstrated the sim-to-real pipeline’s potential for exploring different configurations of visual patterns to improve vision-based reconstruction results. The code and dataset are available at \url{https://github.com/DeepSoRo/DeepSoRoSim2Real}.


\end{abstract}

%% file: 1_intro.tex
\section{Introduction}
Pneumatic soft robots present many unique advantages in difficult manipulation tasks. The high degrees-of-freedom and inherent compliance of soft robots' constituent deformable materials make them safe and reliable in delicate tasks such as harvesting fruit and interacting with deep-sea organisms  \cite{a_l_gunderman_tendon-driven_2022,sinatra_ultragentle_2019}. The high degrees-of-freedom also helps soft robot grippers achieve more versatile and stable grasps with significantly greater contact surface area \cite{a_pagoli_soft_2021,s_puhlmann_rbo_2022}.

However, researchers have struggled with state representation and estimation for soft robots' shapes particularly because of their high degrees of freedom and underlying physical mechanics that are difficult to model \cite{shih_electronic_2020}. Lack of reliable shape estimation or proprioceptive feedback of soft robots severely limits effectiveness and robustness of control strategies and subsequently curtails soft robots' application in real-world applications \cite{george_thuruthel_control_2018}.

To this end, researchers have begun to utilize advances in deep learning to capture the complex and often nonlinear relationship between sensor observation and soft robot state with some success \cite{thuruthel_soft_2019, wang_real-time_2020, randika_estimating_2021}. Training these deep neural network (DNN) models end-to-end requires large and high-quality training data composed of sensor observation and corresponding robot shape pairs.

In contrast to other robotic systems, observing a soft robot's shape is costly. Previously, motion tracking systems have often been used to capture positions of keypoints on a soft robot \cite{thuruthel_soft_2019,toshimitsu_sopra_2021}. However, because these systems rely on tracking separated and discrete marker placements, they suffer from the problem of low-dimensionality and inevitably fail to capture the rich deformation behavior of soft robots. A recent work has used depth-cameras to capture point clouds of soft robots \cite{wang_real-time_2020} and many previous works have used depth-cameras to capture complex deformation behaviors of soft bodies for manipulation \cite{hu_3-d_2019, thach_learning_2022}. However, even for highly controlled settings, a depth-camera can get occluded easily \cite{thach_learning_2022} and the deformation behavior of the soft robot must be confined to prevent occlusion in the camera view. The resulting sample space is constrained by the limitation that no surface of the robot can be covered, meaning no contact with the soft robot could happen during data collection.

\begin{figure}
    \centering
    \includegraphics[width=0.9\linewidth]{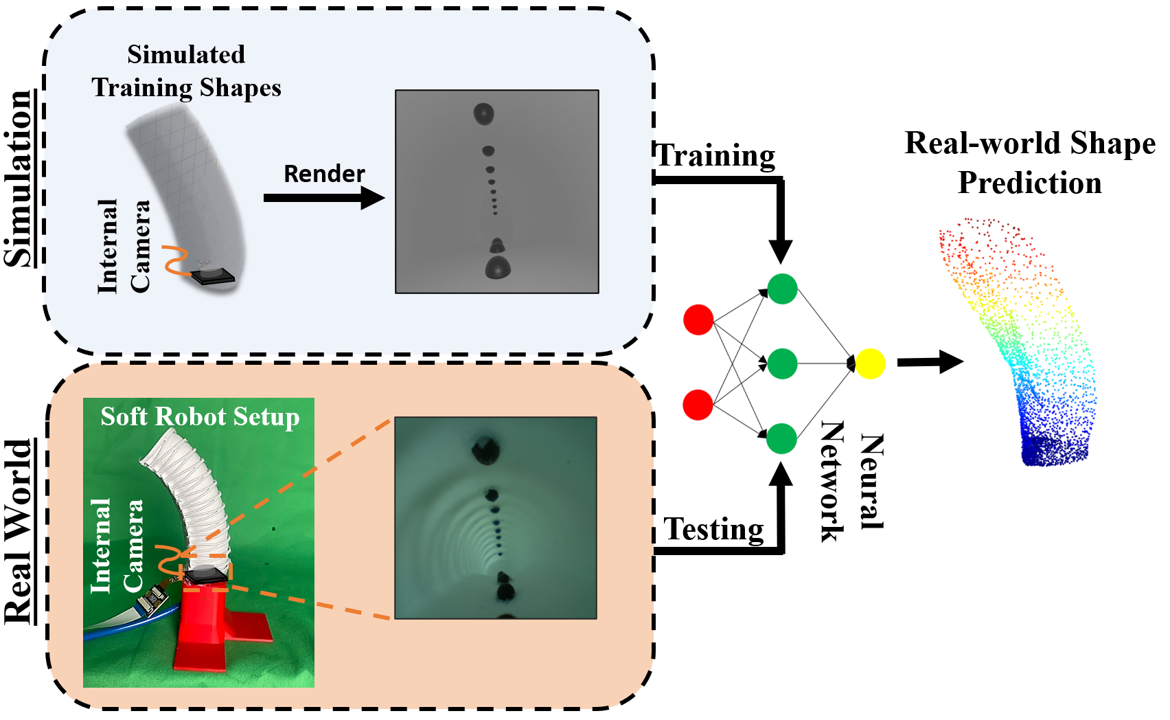}
    \caption{The proposed pipeline of sim-to-real transfer learning for vision-based soft robot. We generated simulation-based point cloud and corresponding internal camera views to train our neural network model. Then, we show that the trained model transfers zero-shot to the real world by testing with real-world images.}
    \label{fig:teaser}
    \vspace{-5mm}
\end{figure}

In other robotics applications that similarly suffer from difficulty in collecting reliable training data in the real world, researchers have successfully taken advantage of simulations to collect the data and transferred the knowledge to the real world in the form of trained deep learning models \cite{si_grasp_2022,pham_pencilnet_2022}. Such sim-to-real transfer learning approaches not only allow data collection with ease but also enable observations under conditions that are physically impossible to observe in the real world. However, to our best knowledge, no previous work has demonstrated sim-to-real transfer learning of soft robot shape reconstruction based on sensory feedback.
In this work, we embedded a camera into a soft robot to get visual feedback of its deformed shape. We placed replicable visual patterns in the pneumatic chamber of the soft robot. We then present and successfully demonstrate a sim-to-real pipeline that allows easy data collection in the simulated environment and high fidelity shape reconstruction based on feedback from a real-world camera embedded in a pneumatic soft robot. We also demonstrated the sim-to-real pipeline's potential to be used as a platform to explore different configurations of visual patterns to improve vision-based reconstruction results. Then the main contributions of the paper can be summarized into three parts:
\begin{enumerate}
    \item Zero-shot sim-to-real transfer learning pipeline for vision-based pneumatic soft robot proprioception sensing with single reference frame scene calibration,
    \item Simulation-driven development and evaluation of visual pattern design heuristics for soft robot propreioception,
    \item Design and evaluation of a novel soft robot with replicable embedded visual feature markers and a wide angle-of-view camera to reduce the sim-to-real gap.
    
\end{enumerate}
Ultimately, this work enables easily implementable high-fidelity soft robot proprioception sensing by dropping the barrier to training data collection and visual feature pattern evaluation through simulation.


\wenzhen{I think the background part of the intro can be further shortened. For the parts that you are going to discuss in details in the related work, you can make it very short in intro}

%% file: 2_prior.tex
\section{Related Work}
\subsection{Soft Robot State Estimation and Proprioception}
Traditionally, soft robot shape estimation relied heavily on simplification through geometric constraints and mechanic theory-based assumptions. By assuming segments of the soft robot can be fitted to a centerline with piece-wise constant curvature (PCC), researchers could deterministically solve for the center-line curve of the robot with reasonable fidelity \cite{polygerinos_modeling_2015, yoo_analytical_2021}. Researchers have also relaxed some of these constraints to scale up the dimensionality of the configuration space and capture more diverse deformation states \cite{y_liu_influence_2021}. However, these methods and models rely heavily on knowing the exact loading conditions on the soft robot which are difficult to sense and are constrained by underlying assumptions on soft robot's mechanics.

Measuring local strain in soft robots with soft resistive strain sensors is a popular method of directly measuring deformation \cite{v_wall_multi-task_2019, tapia_makesense_2020}. Other works have proposed to sense bending and change in curvature \cite{zhao_optoelectronically_2016,gerboni_feedback_2017}. Because individual sensing units (e.g., single strain sensor strip) provide low dimensional feedback, multiple of these sensors must be embedded and ultimately rely on aforementioned underlying assumptions on the soft robot's deformations behaviors to reconstruct the soft robot's shape. Additionally, soft sensors are generally difficult to fabricate and require significant room in the soft robots, limiting their usefulness and scalability. 

\begin{figure}
    \centering
    \includegraphics[width=1.0\linewidth]{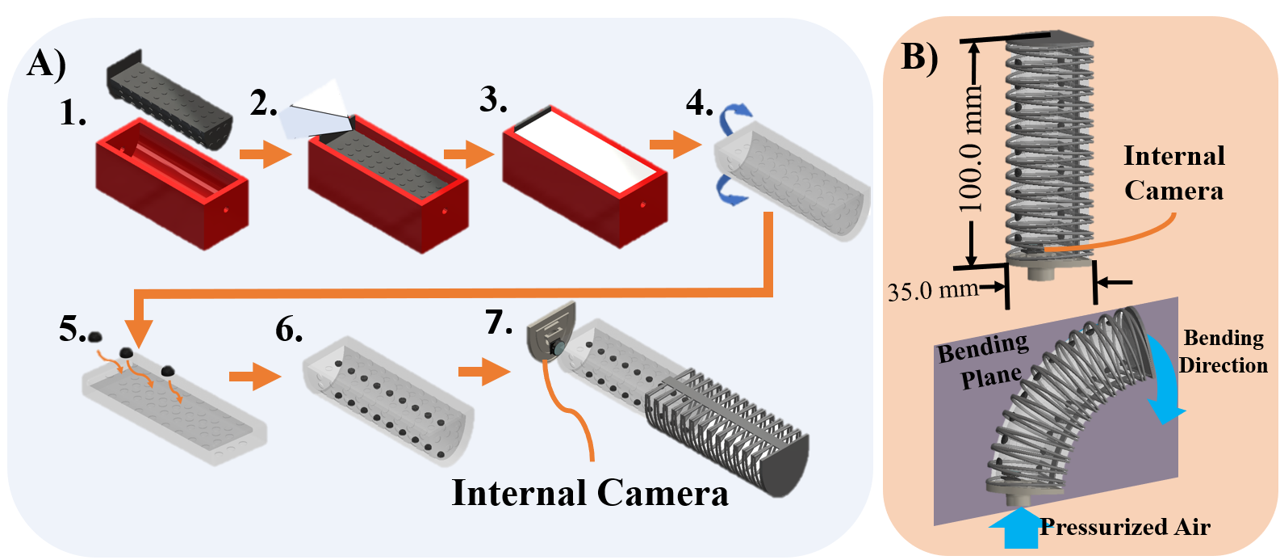}
    \caption{ Fabrication process for the proposed pneumatic soft robot. A1) Assembled mold A2) Uncured elastomer is poured into the mold A3) After curing, the mold is removed A4) The elastomer layer is inverted A5) 3D printed visual patterns are added in the predetermined indented locations of the elastomer A6) The elastomer layer is inverted back to its original state A7) Flexible strain limiting layers and the bottom cap with the wide angle-of-view camera is attached to the elastomer. B) Unactuated and actuated/bent robot configuration with dimensions.}
    \label{fig:fabrication}
    \vspace{-5mm}
\end{figure}
\subsection{Vision-based Soft Robot Proprioception}
Recently, researchers have begun to embed cameras into soft robots to internally observe deformation \cite{hofer_vision-based_2021, she_exoskeleton-covered_2020, wang_real-time_2020}. Because the camera observes deformation indirectly by tracking visual changes, it can be implanted into soft robots with minimal design changes. Furthermore, cameras can provide rich observations of the soft robot's deformation with high spatial precision. Despite this, most previous works severely reduced the soft robot's mechanical degrees of freedom in order to simplify the configuration space. For instance, \cite{she_exoskeleton-covered_2020} models the soft robot's configuration as a 7 degrees-of-freedom series link-and-joint system while \cite{hofer_vision-based_2021} reduced the soft robot further into a 2 degrees-of-freedom ball-and-socket system. 
In this process, the soft robots' unique ability to deform arbitrarily to the contacting environment and object was lost, presenting a constraint on their advantage as well. 

Only \cite{wang_real-time_2020} largely preserves soft robot's high degrees-of-freedom for vision-based proprioception by training a DNN model to map from visual feedback from an embedded camera to the point cloud representation of the deformed soft robot. However, because the deformed robot's shape is observed with a depth-camera in the real world, the training data could only be collected in conditions where the robot is not severely occluded from the depth-camera. Notably, no configurations where the soft robot is in contact with another object could be sampled. The previous work also only could output a partial point cloud of the robot due to the occlusion in the training data \cite{wang_real-time_2020}. Additionally, the data collection process required a highly controlled environment to be setup, adding a severe barrier to using such methods.  

\vspace{-0.2cm}
\subsection{Vision Sim-to-Real Transfer Learning}
\wenzhen{you might be able to shorten this subsection}
Robotics researchers in fields such as manipulation and aerial locomotion have utilized simulated environments to train deep learning visual models \cite{pham_pencilnet_2022, lim_real2sim2real_2022, si_grasp_2022}. In such cases, sim-to-real transfer process involves reducing the difference between the simulation-based rendered images and the real-world images through careful simulation scene adjustments based on some reference real-world images \cite{si_grasp_2022, lim_real2sim2real_2022} or by filtering the images to reduce the impact of hard-to-model optical effects in the renderer \cite{pham_pencilnet_2022}. 

%% file: 3_method.tex
\begin{figure}
    \centering
    \includegraphics[width=1\linewidth]{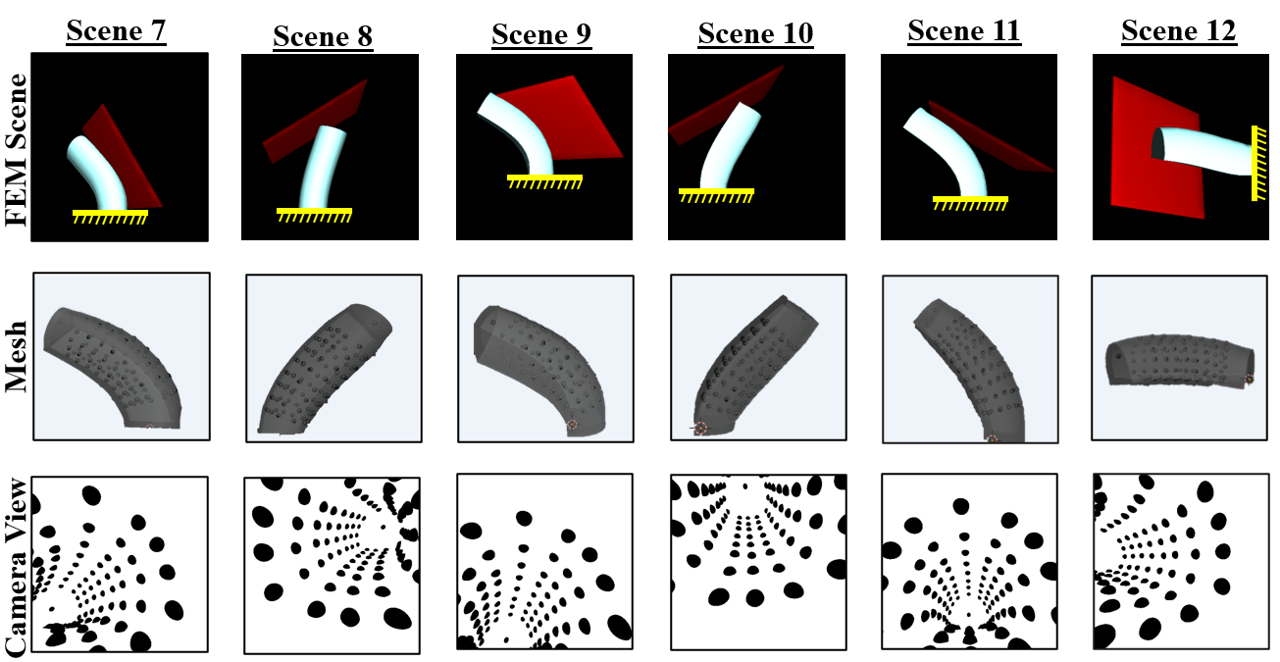}
    \caption{Generated mesh from the FEM simulator and the subsequently rendered internal camera view.}
    \label{fig:sofa}
    \vspace{-5mm}
\end{figure}
   \vspace{-2mm}
\section{Method}


In this section, we present methodologies related to both the proposed sim-to-real pipeline and the design of the soft robot used for evaluation. We also describe the methods implemented to minimize the sim-to-real gap in the generated images and randomization of simulated environments to capture diverse soft robot shape configurations that otherwise would be difficult to capture in the real world.

\subsection{Soft Robot Design and Fabrication}
 \vspace{-1mm}
Pneumatic soft robots generally are composed of two distinct materials \cite{polygerinos_modeling_2015}: the primary elastomer which inflates with increasing internal chamber pressure and the strain-limiting material which enables controlled actuation behavior. To address the aforementioned challenges, we introduce new design requirements to the pneumatic soft robots for vision-based proprioception. 

\wenzhen{I suggest putting this paragraph as a stand-alone sub section to introduce how you design the sensing system: including the camera and the patterns. This should be in addition to the robot design & fabrication}
First, we must embed a wide field-of-view camera into the internal chamber of the robot to observe the internal surface of the soft robot chamber as it deforms. This is a crucial new requirement as these observations feed into the network in inference time to predict the shape of the robot. We must also design visual patterns that can be reliably embedded into the observable internal surface of the soft robot. 

We use Ecoflex 00-30 (Ecoflex 00-30, Smooth-On, Inc.) as the primary elastomer because of its low hysteresis properties with the addition of white pigmentation (Silc Pig, Smooth-On, Inc.) to prevent external lighting conditions from directly affecting the proprioceptive sensing capabilities. As noted in Fig. \ref{fig:fabrication}, we pour the elastomer-pigment mix into the mold that was designed to create precisely positioned craters in the pneumatic cavity of the soft robot. After demolding the elastomer, we embedded 3D-printed 4 mm diameter semi-spherical visual pattern markers into the craters. This procedure allowed the visual patterns to be replicable for both fabrication consistency and simulation scene matching. The new design requirements demand multiple components of different material properties to be seamlessly fused together, which can be difficult, especially for pneumatic soft robots that are prone to leaks and failures. For such cases, 3D printing the strain-limiting layer with flexible resin (Flexible Resin, Formlabs, Inc.) can significantly simplify the assembly and reduce the fabrication quality variability \cite{yoo_analytical_2021}. After embedding the visual pattern markers into the craters, we fused together the elastomer layer, the 3D-printed strain limiting component and the bottom cap with a wide angle-of-view camera (Wide Angle 160° Raspberry Pi Zero Camera Module, Arducam) to get the fully functional pneumatic soft robot with embedded visual pattern markers and a camera.

\begin{figure}
    \centering
    \includegraphics[width=1.0\linewidth]{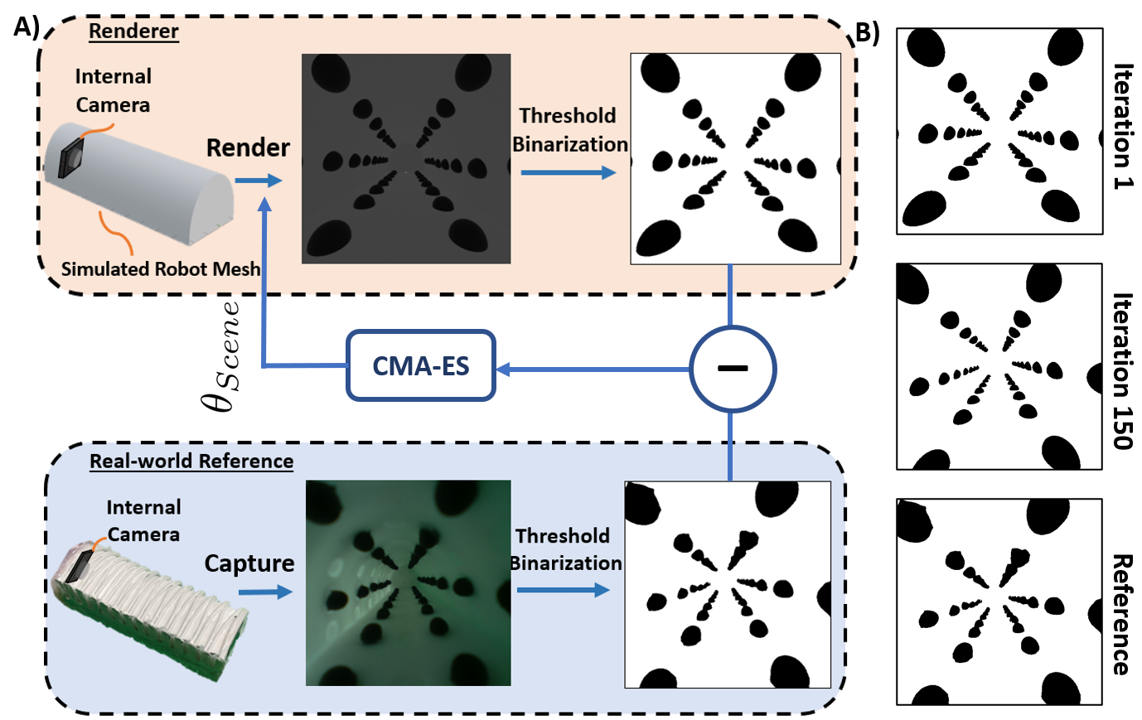}
    \caption{Renderer model calibration process to minimize the sim-to-real gap. A) Covariance matrix adaptation evolution strategy(CMA-ES) is used to iteratively sample and update the renderer scene/model parameters ($\theta_{Scene}$). B)After 20 iterations, the rendered images match the real-world reference image sufficiently well. 
}
    \label{fig:cma}
    \vspace{-5mm}
\end{figure}

\subsection{Soft Robot Numerical Simulation}
\label{SOFA}
We utilize Finite Element Method (FEM)-based numerical simulation to generate physically realistic deformed surface meshes of the developed pneumatic soft robot. And based on the external surface mesh, we extract vertices to construct training point cloud data. To this end, we utilize the open-source SOFA Framework \cite{faure_sofa_2012}. For this paper, SOFA Framework presents many advantages. Specifically, the modular framework, which is largely unique to SOFA, allows us to generate diverse loading conditions with relatively low computational burden. 

We simulate collision conditions in the SOFA scenes to generate diversified deformation data as shown in Fig. \ref{fig:sofa}. We augment FEM-based mesh generation by randomizing three factors: degree of pneumatic actuation of the soft robot, Young's modulus of the inner bending surface material and contact conditions with an external surface. We present the selected bending cases in Fig. \ref{fig:sofa}. We randomized Young's modulus of the soft robot to ensure the generated meshes cover the configuration space of the real-world soft robot.

\subsection{Camera-view Rendering and Renderer Scene Calibration}
\label{BLENDER}

The soft robot design as outlined in the previous sections is easily reproducible and has fewer fabrication steps that are prone to error. The design also significantly simplifies the simulation of the internal-camera view in the renderer scene. We model the visual pattern markers described in Section III B as spherical meshes added in the Blender scene. In scene initialization, the markers are attached to the first undeformed mesh in a scene sequence internal surface mesh nodes in the positions predetermined by the marker placement locations in the real-world soft robot. In the succeeding meshes, the markers track the assigned node movements to simulate embedded visual pattern marker movements in the real world. For each FEM-generated mesh, an image from the camera in the Blender scene is rendered. Because the spherical meshes are added in the renderer component of the pipeline as opposed to being part of the numerical simulator, we are able to easily modify and experiment with different visual pattern marker placements in simulation.   


A significant challenge in simulation-based rendering for sim-to-real transfer learning is accounting for complex lighting and optical effects present in the real world. In this paper, we convert the rendered and real RGB images to binary image representation and eliminate the need to account for hard-to-model optical effects. The intuitive motivation of this process is that by the design of the proposed soft robot and the visual pattern markers, binarization of the images preserves the features we seek to track in the image frame. 

We observed that the fabrication process introduces sim-to-real scene mismatch in the image frame. For example, the camera pose inside the robot could be minutely offset from the designated pose which is not replicated in simulation. To account for this, we iteratively optimize the renderer scene to capture any differences introduced by fabrication error based on covariance matrix adaptation evolution strategy (CMA-ES) \cite{hansen_cma_2016}. CMA-ES is advantageous because it is generally sample efficient and is gradient-free for parameters that we do not have reliable gradient information for.

We capture a reference image of the soft robot in a straight unactuated configuration in the real world. We introduce the scene parameter adjustment vector ($\Delta \theta_{Scene}$) which modifies the scene parameters $\theta_{Scene}$ that we chose to optimize over which takes into account the marker diameter, marker row location in the inner surface, marker distances along a row, camera field of view, and 6 degrees-of-freedom camera pose in the renderer scene for a total of 32 adjustment parameters. We initialize the scene with the ideally designed scene. $\Delta \theta_{Scene}$ is sampled with population $14$ based on heuristics set from the number of parameters \cite{hansen_cma_2016}.  The mean squared difference between the two binary images is then taken and introduced for CMA-ES to update the sampling distribution. After 150 iterations, the scene rendered a close match to the reference image. The process diagram and the resulting rendered images are
featured in Fig. \ref{fig:cma}.

\subsection{Neural Network for 3D Proprioception Sensing}
\label{MODEL}
\vspace{-1mm}
Using the numerical simulation described in Section \ref{SOFA} and the renderer pipeline described in Section \ref{BLENDER}, we can collect rendered images of proprioceptive camera observations and corresponding soft robot shape information under a variety of actuation and loading conditions. Our proposed DNN-based approach encodes visual proprioceptive information to deform a prototype point cloud into the observed state for high-fidelity 3D proprioceptive sensing.

\subsubsection*{\textbf{Model Architecture}}
\begin{figure}
    \centering
    \includegraphics[width=1.0\linewidth, trim={10 120 20 80}, clip]{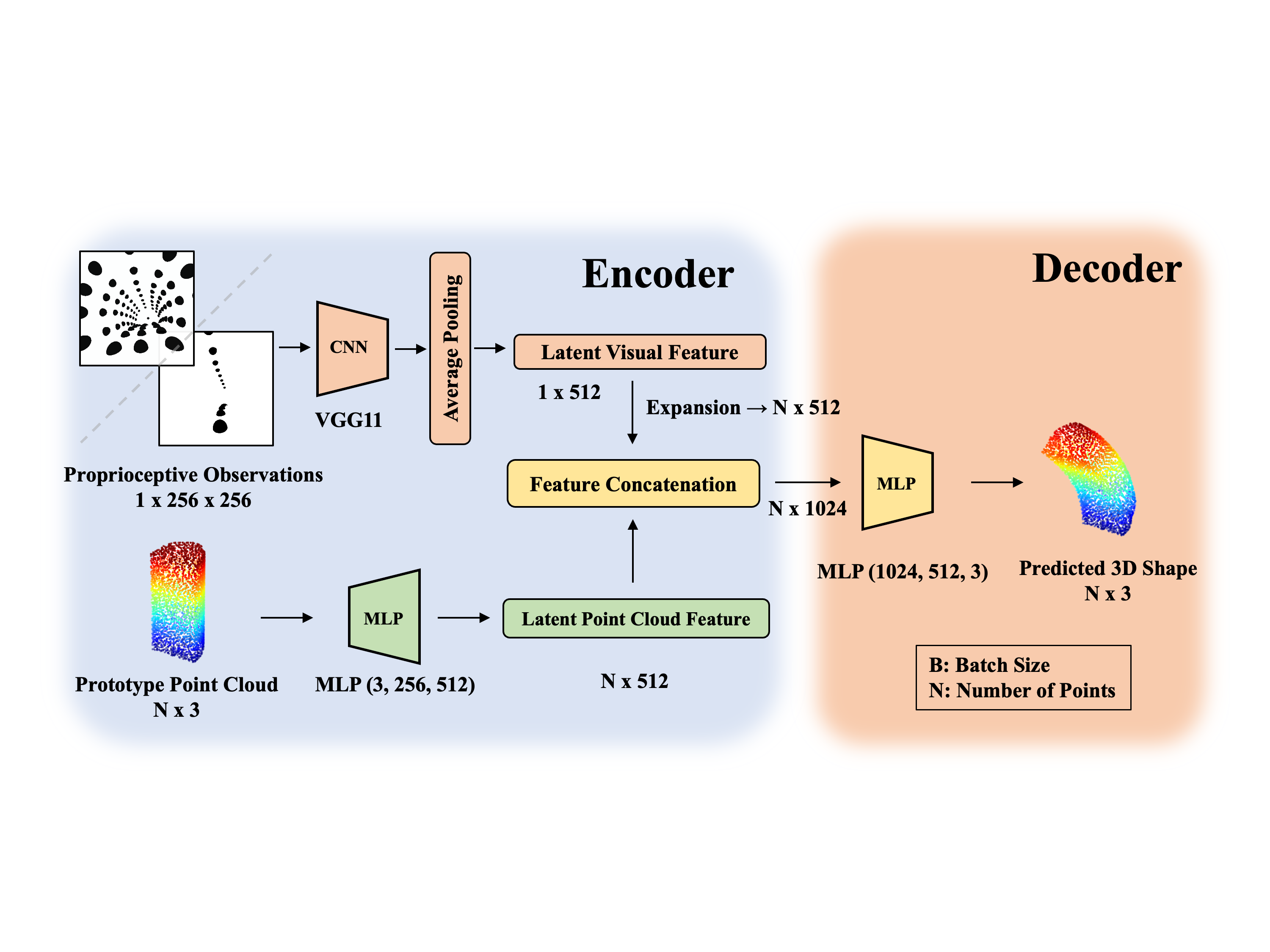}\vspace{-10pt}
    \caption{Deep network architecture based on~\cite{wang_real-time_2020}. The \textbf{Encoder} is a CNN module to encode visual information. The \textbf{Decoder} uses two MLP modules to deform a constant prototype point cloud concatenated with the encoded visual feature to estimate the soft robot's full body shape.}
    \label{fig:model_architecture}
    \vspace{-5mm}
\end{figure}

Our model follows an Encoder-Decoder architecture, as depicted in Fig. \ref{fig:model_architecture}.

\textit{Encoder}.
Our encoder uses a VGG \cite{Simonyan15} module to encode a $1\times256\times256$ binary image as the proprioceptive observation into a $1\times512$ latent visual feature vector. The second dimension of the vector is then expanded by repetition so that the latent visual feature can be concatenated with point cloud features in the Decoder. 

\textit{Decoder}.
Our decoder first uses a two-layer shared MLP (Multilayer Perceptron) with 3-dimensional input and 512-dimensional output to map each point in a prototype point cloud into a high-dimensional latent feature vector. These latent features are then pointwisely concatenated with the latent visual feature, and further mapped back to 3D space as the estimated soft robot 3D shape by another two-layer MLP with 1024-dimensional input and 3-dimensional output. This learnable shape deformation technique was first developed in FoldingNet~\cite{yang2018foldingnet} (2D-to-3D deformation), and then introduced to soft robotics~\cite{wang_real-time_2020} (3D-to-3D deformation).

\subsection{Network Visualization for Marker Placement Design}

\label{marker_placement}
   \vspace{-1mm}
Integrating complex visual feature patterns for soft robot proprioception often requires laborious steps to paint colored patches \cite{wang_real-time_2020} or embed discrete markers one-by-one \cite{hofer_vision-based_2021}. Yet visual feature patterns that are too simple may lead to insufficient visual feedback for accurate proprioception sensing. To discover useful heuristics for handling the trade-off between ease of fabrication and the adequacy of information in the visual feature design for soft robot proprioception, we use network visualization to understand how our model makes predictions based on the observed visual features.

Inspired by Grad-CAM \cite{conf/iccv/SelvarajuCDVPB17}, we back-propagate the Chamfer distance loss and the resulting activation heatmap shows the regions in the image frame that are contributing to the soft robot shape predictions. Using such visualizations, we can understand how our DNN model makes its predictions and provide useful insights for marker placement.

We handcrafted four distinct maker configurations to explore the design space, as seen in Fig. \ref{fig:pattern_overview}. Patterns 1 and 2 are evenly distributed dense patterns, with Pattern 1 including 12 rows of markers and Pattern 2 containing 6 rows. Each row consists of eight semi-spherical markers that are aligned with the longest axis of the soft robot. Pattern 3 has a marker arrangement that is perpendicular to the primary bending plane, as seen in Fig. \ref{fig:fabrication}B, while Pattern 4 includes markers that are aligned with the primary bending plane. We trained on the generated images and corresponding point clouds for each pattern. Then using Grad-CAM, we visualize the activation heatmaps, indicating the model's focus in different regions of the image frame (Fig. \ref{fig:pattern_overview}). From these results, we make hypotheses on effective marker placement strategies.

\textit{Hypothesis (1)}. The model focused on some markers and not others, thus we hypothesize that some patterns with fewer markers can achieve comparable performance to a dense pattern. Hypothesis 1 has an intuitive explanation rooted in the fact that the presented soft robot is an under-actuated system and the markers in the images provide high-dimensional feedback. 

\textit{Hypothesis (2)}. In denser patterns, the markers along the bending surface received more attention (e.g., Pattern 2 in Fig. \ref{fig:pattern_overview}). Thus, we hypothesize that Pattern 4, with its markers aligned on the primary bending plane (Fig. \ref{fig:fabrication}B) would perform better than Pattern 3. This is because the markers along the bending plane would move more in the image frame as the soft robot bends in the preferred direction with lower stiffness. We test these hypotheses in simulation in the following sections.

%% file: 4_exper.tex
 \vspace{-2mm}
\section{Experiments}
   \vspace{-1mm}
\begin{figure}
    \centering
    \includegraphics[width=0.9\linewidth]{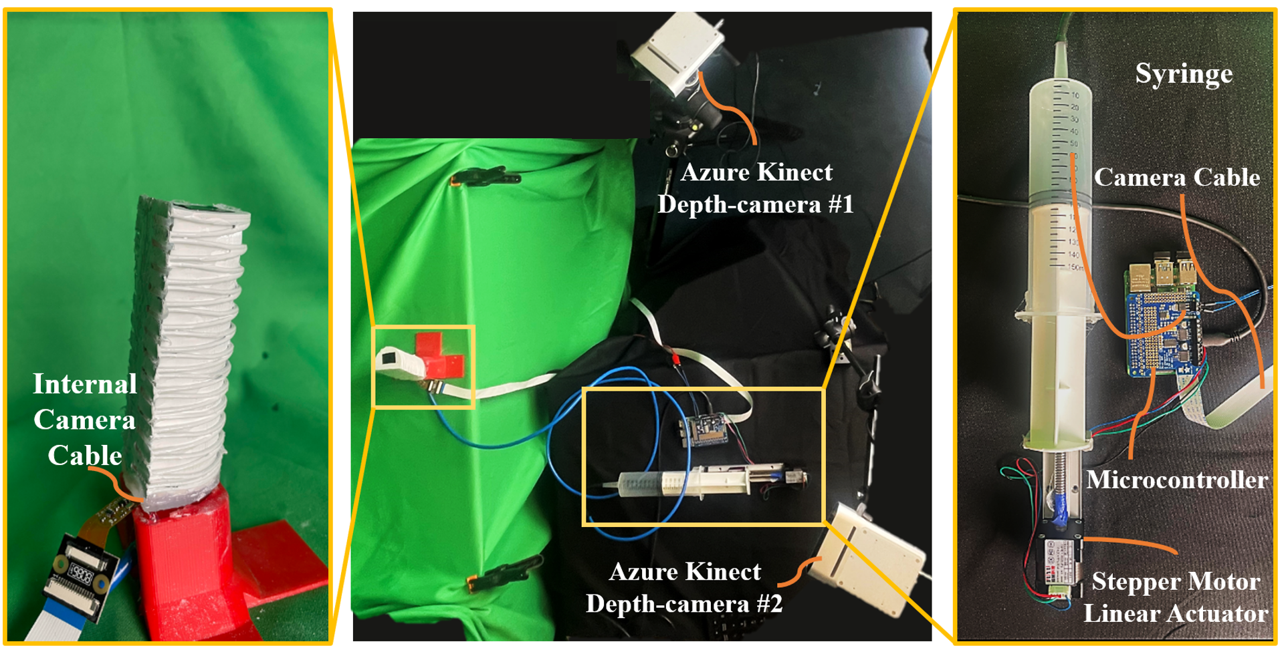}
    \vspace{-2mm}
    \caption{ Experimental setup for the evaluation of the proposed sim-to-real pipeline.}
    \label{fig:setup}
    \vspace{-5mm}
\end{figure}
 \vspace{-1mm}
In this section, we present the experiments performed in simulation and in the real world. First, we discuss our training and evaluation of our proprioceptive DNN model on simulated data. Then, we show that we can use the simulated data to derive useful heuristics for marker pattern design toward improved proprioceptive sensing. We also present our real-world experiment setup that enables us to demonstrate that the model trained on simulated data can zero-shot transfer to the real soft robots.

\subsection{Network Training and Evaluation}
\subsubsection{Data Preparation}
\label{data_prep}
All point clouds were aligned to a single coordinate system. In addition, the point clouds were downsampled to 3174 points to reduce computational cost. As the folding process in our model is a symmetric function that is invariant to input order permutation, downsampling the point cloud would not affect the model's performance.

We simulated 6919 proprioception observations and their associated point clouds from our SOFA and Blender pipeline with 13 distinct simulation scenarios into our dataset. We randomly selected $80\%$ of the data which contains 5535 frames of proprioceptive observations into the training dataset and $20\%$ of the data which contains 1384 frames into the validation dataset across all 36 simulation scenarios.

\subsubsection{Training}
During the training process, both training and testing dataset are separated into mini-batches with \textit{batch\_size} of 50. An Adam \cite{DBLP:journals/corr/KingmaB14} optimizer was defined with initial learning rate of $10^{-4}$, and weight decay coefficient was set to $10^{-6}$. We trained each model with 100 epochs.

\subsubsection{Sim-to-Sim Evaluation}
We evaluated our sim-to-sim model performance with all four handcrafted patterns with the average Chamfer distances from validation dataset were reported in Table. \ref{tab:results}.

\vspace{-2mm}
\subsection{Marker Pattern Design Analysis in Simulation}
The sim-to-sim evaluation results for the four DNN models trained on four visual feature patterns are presented in Table \ref{tab:results}. From the results, we can note that marker pattern 1 which has the most number of markers performed best and the model focused almost on all markers. However, more markers does not always translate to better performance as Pattern 2 under-performed compared to sparser Pattern 3 and 4. These results are consistent with our hypothesis (1).
 
Compared to Pattern 3 with more activated markers, model trained on Pattern 4 concentrated on just the few markers near the robot base. The performance of the model trained on Pattern 4 was also marginally better than that of Pattern 3, indicating that markers on the primary bending plane (Fig.~\ref{fig:fabrication}B) hold more visual information to describe the deformation. Such results provide support for hypothesis (2).

\begin{figure}
    \centering
    \includegraphics[width=0.9\linewidth, trim={30 230 200 20}, clip]{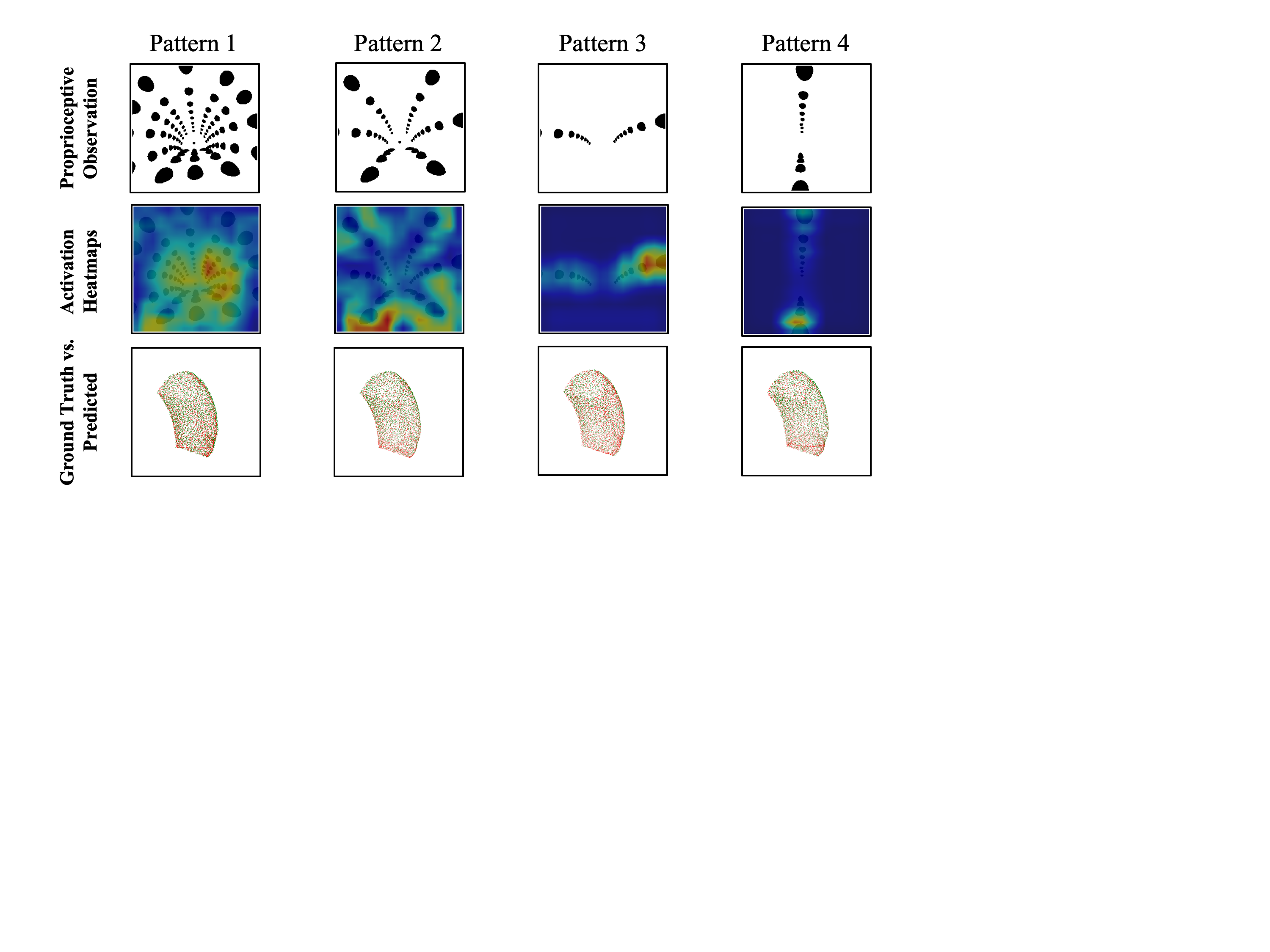}
    \vspace{-5mm}
    \caption{Results of a sim-to-sim evaluation of four distinct pattern designs from our simulation-trained DNN models. Activation heatmaps depict the weighted activation across the image, with red indicating the region where our model has the most attention by back-propagating the Chamfer loss for predicting the shape of the point cloud shown below. 
    Bottom row shows ground truth (red) and predicted (green) point clouds together. In Tab. \ref{tab:results}, the average mean  Chamfer distances for the validation dataset are reported.}
    \label{fig:pattern_overview}
 \vspace{-5mm}
\end{figure}

\subsection{Experiment Setup for the Real-world Soft Robot}\wenzhen{should you make the title something like ``experiment setup for real robot''?}
To evaluate our proprioception sensing method for real robots, we set up a well synchronized system of pneumatic soft robot actuation, groundtruth pointcloud capture, and proprioceptive camera feed, shown in Fig. \ref{fig:setup}. Specifically, for actuation, we sought to minimize any noisy pneumatic pressure changes (e.g., from pump motor inertia) to prevent unsynced mismatch between the depth-cameras and internal camera observations. To this end, we use a stepper motor-actuated syringe pump assembled with a linear actuator (FSL30, FUYU) and a 150 mL syringe. Raspberry Pi 4 (4 Model B, Raspberry) is used to both control the syringe pump through stepper motor controller (Motor HAT, Adafruit) powered by an external variable power supply (Regulated Bench Power Supply, Eventek) and to capture images. The motorized syringe actuate the robot with 5 $cm^3$ of pressurized air increments and external perturbation was applied to bend the soft robot in four directions perpendicular and along the bending plane (Fig. \ref{fig:fabrication}B).
To capture the groud-truth shape of the soft robot, we set up two Azure Kinect DK RGBD cameras at an angle.
We then use Iterative closest point (ICP) algorithm to merge the point clouds from two cameras to get a wider coverage of the robot's shape.
We captured 100 pairs of ground-truth point cloud and proprioceptive camera images for each soft robot. In the first frame of each sequence, we applied ICP to align predicted and ground-truth point clouds at initialization.

\begin{figure}
    \centering
    \includegraphics[width=0.9\linewidth]{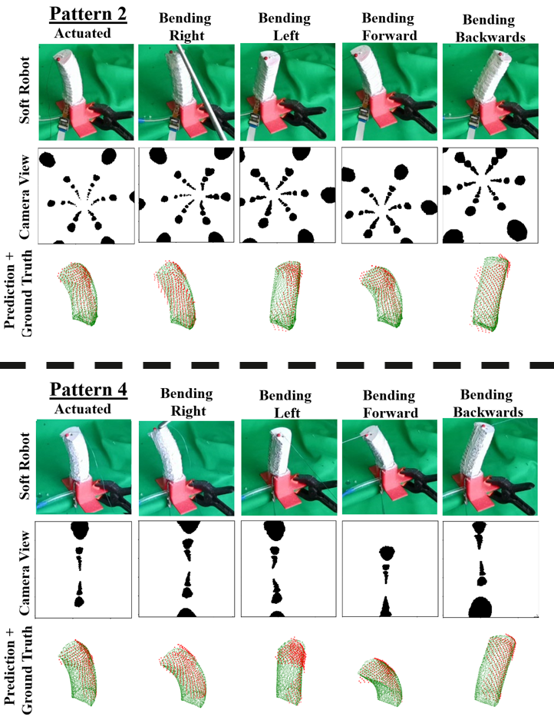}\vspace{-12pt}
    \caption{Comparison between the estimated (green) and observed (red) point cloud shapes of the robot based on the proprioceptive camera images with visual feature pattern 2 and 4. }
    \label{fig:sim2real}
    \vspace{-5mm}
\end{figure}


\vspace{-2mm}
\subsection{Zero-shot Sim-to-real Transfer Evaluation}
We fabricated the proposed pneumatic soft robot with Pattern 2 and Pattern 4 from Fig. \ref{fig:pattern_overview}\wenzhen{don't refer to figures in appendix} to verify our pipeline's sim-to-real performance. Namely, we seeked to test the following hypotheses in these experiments: 1. the presented simulation calibration enables zero-shot sim-to-real transfer of the model trained solely on simulated data 2. Pattern 4 with sparser markers in the vertical image frame performs similarly to Pattern 2 with denser markers. Then we evaluate the two soft robots on five distinct bending behaviors that the proposed soft robot can experience: 1. actuated bending as shown in Fig. \ref{fig:fabrication}B,\wenzhen{wrong figure pointer?} 2. bending forward, 3. bending backward 4,. bending right, and 5. bending left. We compared the predicted shape of the soft robot against the real-world shape obtained by the experiment setup presented in Fig. \ref{fig:setup}\wenzhen{do you mean Table?}.

We evaluate our method's performance with two metrics: mean averaged unidirectional Chamfer distance and tip position mean average error (MAE). The mean averaged unidirectional Chamfer distance provides the average distance from the partial real-world point cloud to the nearest points in the learned model output point cloud while the tip position MAE is calculated by finding the average distance between the tip position in the observed point cloud to the tip position of the predicted shape. While the unidirectional Chamfer distance compares the geometric differences between the observed and the predicted overall shapes of the soft robot, tip position MAE provides a more intuitive single point-wise deviation error between the two predicted shapes.

\begin{table}
\begin{threeparttable}
\caption{Simulation-trained Model Evaluation in Simulation and Real World}
\label{tab:results}
\setlength\tabcolsep{0pt} 
\begin{tabular*}{\columnwidth}{@{\extracolsep{\fill}} ll ccc}
\toprule
    Test Data Source & Marker Pattern & 
     \multicolumn{2}{c}{Performance Metrics [mm]} \\ 

     & &Chamfer Distance $\downarrow$ & Tip MAE $\downarrow$\\
\midrule
     Simulation & Pattern 1 [96 markers] & 2.63 &  1.97\\
\addlinespace
     Simulation & Pattern 2 [48 markers] & 3.67 & 1.86\\
\addlinespace
    Simulation & Pattern 3 [16 markers] & 3.31 & 4.25\\
\addlinespace
     Simulation & Pattern 4 [16 markers] & 3.25 & 3.16\\
\addlinespace
    Real World & Pattern 2  [48 markers]& 8.85 & 10.12\\
\addlinespace
     Real World & Pattern 4  [16 markers]&10.56 & 13.78\\
\bottomrule
\end{tabular*}
\smallskip

\scriptsize
  \vspace{-8pt}       
\wenzhen{You need to improve the format of this table.}
\wenzhen{Why don't you have the tip measurement in simulation? I suppose that should be easy to get}
\end{threeparttable}
\vspace{-5mm}
\end{table}

Overall, both patterns performed notably well in transferring the simulation-trained model to the real world as it can be noted by the qualitative visualizations of the results in Fig. \ref{fig:sim2real}. We may also note that the error was swayed heavily by outliers. For Pattern 4, there is a significantly larger prediction error when bending left. The exceptional performance of Pattern 2 model in capturing the left bending case indicates this failure may be because Pattern 4 is not optimal for capturing deformation off of the primary bending plane. We observed that with the left and right bending frames removed, Pattern 4 average Chamfer distance decreased significantly to $6.80$ mm while the tip position MAE decreased to $7.52$ mm. A significant portion of the remaining error may be due to the uneven surface of the soft robot and noise and artifacts from the depth cameras. Our method's tip position MAE is similar to those previously reported in literature with similar soft robot dimensions \cite{thuruthel_soft_2019}. However, our method outputs a high-dimensional representation (3174 points in 3D space) estimate of the soft robot's shape to uniquely allow high-fidelity 3D reconstruction of the soft robot even in arbitrary nonplanar bending and external loading conditions.

%% file: 5_disc.tex
 \vspace{-2mm}
\section{Conclusions}
 \vspace{-2mm}
In this work, we propose and validate the first sim-to-real pipeline for vision-based proprioception of soft robots. Single real-world reference image is used to calibrate the renderer scene and effectively close the sim-to-real gap, enabling our demonstrated ability to zero-shot transfer our simulation-trained model to the real-world data. We also explore the possible application of the simulation pipeline for visual pattern marker optimization by testing our hypotheses in simulation and showing the real-world results seem to validate the trends observed in simulation. The process in this work, however, was limited to ad hoc exploration of the pattern design space. For our future work, we plan to replace the renderer in our current sim-to-real pipeline with a differentiable renderer which can differentiably relate the 2D images back to the 3D scenes. The resulting pipeline will enable us to backpropogate in training time to optimize the visual feature patterns for vision-based proprioception.

%% file: 6_ackn.tex
%


%% file: output.bbl
\begin{thebibliography}{10}

\bibitem{a_l_gunderman_tendon-driven_2022}
{A. L. Gunderman}, {J. A. Collins}, {A. L. Myers}, {R. T. Threlfall}, and {Y.
  Chen}, ``Tendon-{Driven} {Soft} {Robotic} {Gripper} for {Blackberry}
  {Harvesting},'' {\em IEEE Robotics and Automation Letters}, vol.~7,
  pp.~2652--2659, Apr. 2022.

\bibitem{sinatra_ultragentle_2019}
N.~R. Sinatra, C.~B. Teeple, D.~M. Vogt, K.~K. Parker, D.~F. Gruber, and R.~J.
  Wood, ``Ultragentle manipulation of delicate structures using a soft robotic
  gripper,'' {\em Science Robotics}, vol.~4, p.~eaax5425, Aug. 2019.
\newblock Publisher: American Association for the Advancement of Science.

\bibitem{a_pagoli_soft_2021}
{A. Pagoli}, {F. Chapelle}, {J. A. Corrales}, {Y. Mezouar}, and {Y. Lapusta},
  ``A {Soft} {Robotic} {Gripper} {With} an {Active} {Palm} and {Reconfigurable}
  {Fingers} for {Fully} {Dexterous} {In}-{Hand} {Manipulation} *,'' {\em IEEE
  Robotics and Automation Letters}, vol.~6, pp.~7706--7713, Oct. 2021.

\bibitem{s_puhlmann_rbo_2022}
{S. Puhlmann}, {J. Harris}, and {O. Brock}, ``{RBO} {Hand} 3: {A} {Platform}
  for {Soft} {Dexterous} {Manipulation},'' {\em IEEE Transactions on Robotics},
  pp.~1--16, 2022.

\bibitem{shih_electronic_2020}
B.~Shih, D.~Shah, J.~Li, T.~G. Thuruthel, Y.-L. Park, F.~Iida, Z.~Bao,
  R.~Kramer-Bottiglio, and M.~T. Tolley, ``Electronic skins and machine
  learning for intelligent soft robots,'' {\em Science Robotics}, vol.~5,
  p.~eaaz9239, Apr. 2020.
\newblock Publisher: American Association for the Advancement of Science.

\bibitem{george_thuruthel_control_2018}
T.~George~Thuruthel, Y.~Ansari, E.~Falotico, and C.~Laschi, ``Control
  {Strategies} for {Soft} {Robotic} {Manipulators}: {A} {Survey},'' {\em Soft
  Robotics}, vol.~5, pp.~149--163, Apr. 2018.
\newblock Publisher: Mary Ann Liebert, Inc., publishers.

\bibitem{thuruthel_soft_2019}
T.~G. Thuruthel, B.~Shih, C.~Laschi, and M.~T. Tolley, ``Soft robot perception
  using embedded soft sensors and recurrent neural networks,'' {\em Science
  Robotics}, vol.~4, p.~eaav1488, Jan. 2019.

\bibitem{wang_real-time_2020}
R.~Wang, S.~Wang, S.~Du, E.~Xiao, W.~Yuan, and C.~Feng, ``Real-{Time} {Soft}
  {Body} {3D} {Proprioception} via {Deep} {Vision}-{Based} {Sensing},'' {\em
  IEEE ROBOTICS AND AUTOMATION LETTERS}, vol.~5, no.~2, p.~8, 2020.

\bibitem{randika_estimating_2021}
K.~Randika and K.~Takemura, ``Estimating the {Shape} of {Soft} {Pneumatic}
  {Actuators} using {Active} {Vibroacoustic} {Sensing},'' in {\em 2021
  {IEEE}/{RSJ} {International} {Conference} on {Intelligent} {Robots} and
  {Systems} ({IROS})}, (Prague, Czech Republic), pp.~7189--7194, IEEE, Sept.
  2021.

\bibitem{toshimitsu_sopra_2021}
Y.~Toshimitsu, K.~W. Wong, T.~Buchner, and R.~Katzschmann, ``{SoPrA}:
  {Fabrication} \& {Dynamical} {Modeling} of a {Scalable} {Soft} {Continuum}
  {Robotic} {Arm} with {Integrated} {Proprioceptive} {Sensing},'' in {\em 2021
  {IEEE}/{RSJ} {International} {Conference} on {Intelligent} {Robots} and
  {Systems} ({IROS})}, (Prague, Czech Republic), pp.~653--660, IEEE, Sept.
  2021.

\bibitem{hu_3-d_2019}
Z.~Hu, T.~Han, P.~Sun, J.~Pan, and D.~Manocha, ``3-{D} {Deformable} {Object}
  {Manipulation} {Using} {Deep} {Neural} {Networks},'' {\em IEEE ROBOTICS AND
  AUTOMATION LETTERS}, vol.~4, no.~4, p.~7, 2019.

\bibitem{thach_learning_2022}
B.~Thach, B.~Y. Cho, A.~Kuntz, and T.~Hermans, ``Learning {Visual} {Shape}
  {Control} of {Novel} {3D} {Deformable} {Objects} from {Partial}-{View}
  {Point} {Clouds},'' in {\em 2022 {International} {Conference} on {Robotics}
  and {Automation} ({ICRA})}, pp.~8274--8281, IEEE Press, 2022.
\newblock event-place: Philadelphia, PA, USA.

\bibitem{si_grasp_2022}
Z.~Si, Z.~Zhu, A.~Agarwal, S.~Anderson, and W.~Yuan, ``Grasp {Stability}
  {Prediction} with {Sim}-to-{Real} {Transfer} from {Tactile} {Sensing},'' Aug.
  2022.
\newblock arXiv:2208.02885 [cs].

\bibitem{pham_pencilnet_2022}
H.~X. Pham, A.~Sarabakha, M.~Odnoshyvkin, and E.~Kayacan, ``{PencilNet}:
  {Zero}-{Shot} {Sim}-to-{Real} {Transfer} {Learning} for {Robust} {Gate}
  {Perception} in {Autonomous} {Drone} {Racing},'' July 2022.
\newblock arXiv:2207.14131 [cs].

\bibitem{polygerinos_modeling_2015}
P.~Polygerinos, Z.~Wang, J.~T.~B. Overvelde, K.~C. Galloway, R.~J. Wood,
  K.~Bertoldi, and C.~J. Walsh, ``Modeling of {Soft} {Fiber}-{Reinforced}
  {Bending} {Actuators},'' {\em IEEE Transactions on Robotics}, vol.~31,
  pp.~778--789, June 2015.

\bibitem{yoo_analytical_2021}
U.~Yoo, Y.~Liu, A.~D. Deshpande, and F.~Alamabeigi, ``Analytical {Design} of a
  {Pneumatic} {Elastomer} {Robot} {With} {Deterministically} {Adjusted}
  {Stiffness}.,'' {\em IEEE Robotics Autom. Lett.}, vol.~6, no.~4,
  pp.~7781--7788, 2021.

\bibitem{y_liu_influence_2021}
{Y. Liu}, {U. Yoo}, {S. Ha}, {S. F. Atashzar}, and {F. Alambeigi}, ``Influence
  of {Antagonistic} {Tensions} on {Distributed} {Friction} {Forces} of
  {Multisegment} {Tendon}-{Driven} {Continuum} {Manipulators} {With}
  {Irregular} {Geometry},'' {\em IEEE/ASME Transactions on Mechatronics},
  pp.~1--11, 2021.

\bibitem{v_wall_multi-task_2019}
{V. Wall} and {O. Brock}, ``Multi-{Task} {Sensorization} of {Soft} {Actuators}
  {Using} {Prior} {Knowledge},'' in {\em 2019 {International} {Conference} on
  {Robotics} and {Automation} ({ICRA})}, pp.~9416--9421, May 2019.
\newblock Journal Abbreviation: 2019 International Conference on Robotics and
  Automation (ICRA).

\bibitem{tapia_makesense_2020}
J.~Tapia, E.~Knoop, M.~Mutný, M.~A. Otaduy, and M.~Bächer, ``{MakeSense}:
  {Automated} {Sensor} {Design} for {Proprioceptive} {Soft} {Robots},'' {\em
  Soft Robotics}, vol.~7, pp.~332--345, June 2020.
\newblock Publisher: Mary Ann Liebert, Inc., publishers.

\bibitem{zhao_optoelectronically_2016}
H.~Zhao, K.~O’Brien, S.~Li, and R.~F. Shepherd, ``Optoelectronically
  innervated soft prosthetic hand via stretchable optical waveguides,'' {\em
  Science Robotics}, vol.~1, p.~eaai7529, Dec. 2016.

\bibitem{gerboni_feedback_2017}
G.~Gerboni, A.~Diodato, G.~Ciuti, M.~Cianchetti, and A.~Menciassi, ``Feedback
  {Control} of {Soft} {Robot} {Actuators} via {Commercial} {Flex} {Bend}
  {Sensors},'' {\em IEEE/ASME Transactions on Mechatronics}, vol.~22,
  pp.~1881--1888, Aug. 2017.

\bibitem{hofer_vision-based_2021}
M.~Hofer, C.~Sferrazza, and R.~D’Andrea, ``A {Vision}-{Based} {Sensing}
  {Approach} for a {Spherical} {Soft} {Robotic} {Arm},'' {\em Frontiers in
  Robotics and AI}, vol.~8, p.~630935, Feb. 2021.

\bibitem{she_exoskeleton-covered_2020}
Y.~She, S.~Q. Liu, P.~Yu, and E.~Adelson, ``Exoskeleton-covered soft finger
  with vision-based proprioception and tactile sensing,'' in {\em 2020 {IEEE}
  {International} {Conference} on {Robotics} and {Automation} ({ICRA})},
  (Paris, France), pp.~10075--10081, IEEE, May 2020.

\bibitem{lim_real2sim2real_2022}
V.~Lim, H.~Huang, L.~Y. Chen, J.~Wang, J.~Ichnowski, D.~Seita, M.~Laskey, and
  K.~Goldberg, ``{Real2Sim2Real}: {Self}-{Supervised} {Learning} of {Physical}
  {Single}-{Step} {Dynamic} {Actions} for {Planar} {Robot} {Casting},'' in {\em
  2022 {International} {Conference} on {Robotics} and {Automation} ({ICRA})},
  (Philadelphia, PA, USA), pp.~8282--8289, IEEE, May 2022.

\bibitem{faure_sofa_2012}
F.~Faure, C.~Duriez, H.~Delingette, J.~Allard, B.~Gilles, S.~Marchesseau,
  H.~Talbot, H.~Courtecuisse, G.~Bousquet, I.~Peterlik, and S.~Cotin, ``{SOFA}:
  {A} {Multi}-{Model} {Framework} for {Interactive} {Physical} {Simulation},''
  in {\em Soft {Tissue} {Biomechanical} {Modeling} for {Computer} {Assisted}
  {Surgery}} (Y.~Payan, ed.), pp.~283--321, Berlin, Heidelberg: Springer Berlin
  Heidelberg, 2012.

\bibitem{hansen_cma_2016}
N.~Hansen, ``The {CMA} {Evolution} {Strategy}: {A} {Tutorial},'' 2016.

\bibitem{Simonyan15}
K.~Simonyan and A.~Zisserman, ``Very deep convolutional networks for
  large-scale image recognition,'' in {\em International Conference on Learning
  Representations}, 2015.

\bibitem{yang2018foldingnet}
Y.~Yang, C.~Feng, Y.~Shen, and D.~Tian, ``Foldingnet: Point cloud auto-encoder
  via deep grid deformation,'' in {\em Proceedings of the IEEE Conference on
  Computer Vision and Pattern Recognition}, pp.~206--215, 2018.

\bibitem{conf/iccv/SelvarajuCDVPB17}
R.~R. Selvaraju, M.~Cogswell, A.~Das, R.~Vedantam, D.~Parikh, and D.~Batra,
  ``Grad-cam: Visual explanations from deep networks via gradient-based
  localization.,'' in {\em ICCV}, pp.~618--626, IEEE Computer Society, 2017.

\bibitem{DBLP:journals/corr/KingmaB14}
D.~P. Kingma and J.~Ba, ``Adam: {A} method for stochastic optimization,'' in
  {\em 3rd International Conference on Learning Representations, {ICLR} 2015,
  San Diego, CA, USA, May 7-9, 2015, Conference Track Proceedings} (Y.~Bengio
  and Y.~LeCun, eds.), 2015.

\end{thebibliography}
